\documentclass[sigconf]{acmart}

\def\BibTeX{{\rm B\kern-.05em{\sc i\kern-.025em b}\kern-.08emT\kern-.1667em\lower.7ex\hbox{E}\kern-.125emX}}
    
\copyrightyear{2019}
\acmYear{2019}
\setcopyright{iw3c2w3}
\acmConference[Report '19]{Report}{2019}{Berlin, DE}
\usepackage{amsmath,amssymb,amsfonts}
\usepackage{algorithmic}
\usepackage{graphicx}
\usepackage{textcomp}
\usepackage{diagbox}

\newcommand*{\origrightarrow}{}

\renewcommand*{\textrightarrow}{\fontfamily{cmr}\selectfont\origrightarrow}

\DeclareMathOperator{\tf}{tf}
\DeclareMathOperator{\idf}{idf}

\usepackage{amsthm}
\theoremstyle{definition}
\newtheorem{definition}{Definition}[section]

\usepackage{times,amsmath,epsfig}
\usepackage{float}
\usepackage{pgf}
\usepackage{pifont}
\usepackage{pgfplots}
\usepgfplotslibrary{groupplots}
\usepackage{subfigure}
\usepackage{graphicx}

\usepackage{caption}
\usepackage{multirow}
\usepackage{makecell}

\usepackage{enumitem}
\usepackage{amsmath}
\usepackage{environ}
\usepackage{verbatim}
\usepackage[T1]{fontenc}
\usepackage{adjustbox}
\usepackage{comment}
\usetikzlibrary{fit,calc}
\usepgfplotslibrary{external}

\definecolor{r1}{RGB}{228, 26, 28}
\definecolor{r2}{RGB}{55, 126, 184}
\definecolor{r3}{RGB}{77, 175, 74}
\definecolor{r4}{RGB}{152, 78, 163}
\definecolor{r5}{RGB}{255, 127, 0}
\definecolor{r6}{RGB}{166,86,40}
\definecolor{r7}{RGB}{0,0,0}

\pgfplotscreateplotcyclelist{mycolorlist}{%
	r1,mark=none\\%
	r2,mark=none\\%
	r3,mark=none\\%
	r4,mark=none\\%
	r5,mark=none\\%
	r6,mark=none\\%
	r7,mark=none\\%
}

\usepackage{color}
\usepackage{tikz}
\usetikzlibrary{arrows.meta}

\usepackage{array}
\newcolumntype{?}{!{\vrule width 1pt}}

\usepackage{ctable}

\usepackage{amsfonts}
\usepackage{amsmath}
\usepackage{resizegather}
\usepackage{url}
\usepackage{tkz-euclide}
\usepackage{colortbl}
\usepackage{xcolor}
\newcolumntype{e}{>{\columncolor{red!10}}r}
\newcolumntype{L}{@{}>{\kern\tabcolsep}l<{\kern\tabcolsep}}
\newcolumntype{R}{@{}>{\kern\tabcolsep}r<{\kern\tabcolsep}}

\begin{document}

\setlength{\abovedisplayskip}{3pt}
\setlength{\belowdisplayskip}{3pt}

%
% The "title" command has an optional parameter, allowing the author to define a "short title" to be used in page headers.
\title{\system{}: Two-stage Active Learning for Error Detection - Technical Report}

%
% The "author" command and its associated commands are used to define the authors and their affiliations.
% Of note is the shared affiliation of the first two authors, and the "authornote" and "authornotemark" commands
% used to denote shared contribution to the research.

\author{Felix Neutatz}
\email{felix.neutatz@dfki.de}
\affiliation{%
  \institution{DFKI GmbH}
}
\author{Mohammad Mahdavi}
\email{mahdavilahijani@tu-berlin.de}
\affiliation{%
  \institution{TU Berlin}
}
\author{Ziawasch Abedjan}
\email{abedjan@tu-berlin.de}
\affiliation{%
  \institution{TU Berlin}
}

%
% By default, the full list of authors will be used in the page headers. Often, this list is too long, and will overlap
% other information printed in the page headers. This command allows the author to define a more concise list
% of authors' names for this purpose.
%\renewcommand{\shortauthors}{Neutatz et al.}

\newcommand{\za}[1]{\textcolor{blue}{Ziawasch: #1}}
\newcommand{\mohammad}[1]{\textcolor{red}{Mohammad: #1}}
\newcommand{\felix}[1]{\textcolor{purple}{Felix: #1}}
\newcommand{\imagescale}{0.68}

\newcommand{\limitflights}{120}
\newcommand{\limitmovies}{300}
\newcommand{\limitrestaurants}{500}
\newcommand{\limitbeers}{340}
\newcommand{\limitaddress}{150}
\newcommand{\limithospital}{500}

\newcommand{\system}{\textsc{ED2}}

\newcommand*\circled[1]{\tikz[baseline=(char.base)]{
		\node[shape=circle,draw,inner sep=1pt] (char) {\footnotesize #1};}}

%
% The abstract is a short summary of the work to be presented in the article.
\begin{abstract}
	Traditional error detection approaches require user-defined parameters and rules. Thus, the user has to know both the error detection system and the data. 
	However, we can also formulate error detection as a semi-supervised classification problem that only requires domain expertise. 
	The challenges for such an approach are twofold: (1)~to represent the data in a way that enables a classification model to identify various kinds of data errors, and (2)~to pick the most promising data values for learning.
	In this paper, we address these challenges with \system{}, our new example-driven error detection method. 
	First, we present a new two-dimensional multi-classifier sampling strategy for active learning.
	Second, we propose novel multi-column features. 
	The combined application of these techniques provides fast convergence of the classification task with high detection accuracy. 
	On several real-world datasets, \system{} requires, on average, less than 1\%~labels to outperform existing error detection approaches.
	
	This report extends the peer-reviewed paper \emph{ED2: A Case for Active Learning in Error Detection}~\cite{neutatz2019ed2}. All source code related to this project is available on GitHub\footnote{\url{https://github.com/BigDaMa/ExampleDrivenErrorDetection}}.
\end{abstract}

%
% The code below is generated by the tool at http://dl.acm.org/ccs.cfm.
% Please copy and paste the code instead of the example below.
%

%\begin{CCSXML}
%<ccs2012>
%<concept>
%<concept_id>10002951.10002952.10003219.10003218</concept_id>
%<concept_desc>Information systems~Data cleaning</concept_desc>
%<concept_significance>500</concept_significance>
%</concept>
%</ccs2012>
%\end{CCSXML}

%https://dl.acm.org/ccs/ccs.cfm?id=10003218&lid=0.10002951.10002952.10003219.10003218
% Information systems →  Data management systems →  Information integration →  Data cleaning
%\ccsdesc[500]{Information systems~Data cleaning}
%\ccsdesc[300]{Computer systems organization~Redundancy}
%\ccsdesc[100]{Networks~Network reliability}

%
% Keywords. The author(s) should pick words that accurately describe the work being
% presented. Separate the keywords with commas.
%keywords{datasets, neural networks, gaze detection, text tagging}

%
% This command processes the author and affiliation and title information and builds
% the first part of the formatted document.
\maketitle

\section{Introduction}
Data cleaning is an essential process in maintaining high data quality and reliable analytics.
A core step in data cleaning is identifying erroneous values and repairing them. 

In this paper, we consider the problem of error detection. Accurate detection of errors improves the performance of data-driven error correction methods~\cite{rekatsinas2017holoclean} and relieves data engineers from manual verification of false positives. In fact, many companies still rely on manual repair procedures~\cite{abedjan2016detecting}.
Research on error detection has pursued different strategies, such as rule-based~\cite{dallachiesa2013nadeef}, quantitative~\cite{pit2016outlier}, pattern-based~\cite{kandel2012enterprise}, or dictionary-based~\cite{chu2015katara} methods.
%parameters are bad
However, all these methods require either qualitative parameters, such as rules, or quantitative parameters, such as outlier thresholds.
%examples for why parameters are bad
Identifying promising outlier thresholds requires the user to search and evaluate a wide range of statistical parameters. % hyperparameter optimization~\cite{pit2016outlier}.
Similarly, formulating rules requires rudimentary programming knowledge, e.g., to formulate regular expressions or functional dependencies.
Yet, users often are unaware of the pre-existing set of rules and have to undergo time-consuming browsing and exploration routines to identify applicable cleaning rules.
%Therefore, most of the time, the person in charge creates purpose-built cleaning scripts or uses interactive tools, such as Excel, OpenRefine, or Trifacta~\cite{kandel2012enterprise}, to clean their datasets. 
%In fact, the users have to undergo time-consuming browsing and exploration routines to identify applicable cleaning rules. 

A different approach to error detection that does not require the user to have prior tool-specific knowledge or spend time wrangling with the data is to consider it as a classification task. 
The idea is to have the user label a subset of the dataset as erroneous or correct. Then, classification models learn to label the rest of the dataset accordingly. This approach requires only domain knowledge from the users and unburdens them from parameter tuning and rule declaration.

%can find correlation that the user is not aware of
%An advantage of such a data-driven method is that it can exploit hidden correlations that the user was unaware of to detect errors. For instance, in one of our experimental datasets that contains flight information from various sources, our method exploits a hidden correlation between source identifiers and erroneous values. Thus, our method learns that some sources are more trustworthy than others. \mohammad{not sure about mentioning this "an advantage": such a approach could have different advantages and this one is not the most important ones. furthemore, here we still are talking about the whole family of classification-based approaches (not our particular \system{}).}

ActiveClean~\cite{krishnan2016activeclean} and BoostClean~\cite{krishnan2017boostclean} are two recent approaches that apply error classification for tuples. 
Both approaches focus on identifying and cleaning erroneous tuples based on their impact on an application-specific machine learning task, such as house price prediction. The limitation of these approaches is twofold: First, their sampling strategy for obtaining labels depends on an application-specific machine learning task. In the absence of such a task, one has to resort to a more general sampling strategy. Second, their approach and feature selection are designed to detect erroneous tuples. 
In many application scenarios, identifying only the tuple is not enough. One has to find the exact position of an actual data error, namely the actual data cells. 
To formulate a classification task for this goal, the following two challenges emerge:

\begin{table}[t!]
	\centering
	\caption{Sample dataset with errors.} %Colored background specifies that there is a user label available for the corresponding cell. An erroneous label is red and a correct label is green. If a cell is erroneous but no label is available, the text is colored red.}
	\resizebox{\columnwidth}{!}{
	\begin{tabular}{|l|r|r|r|}  \hline
        Position & Salary & Phone & Company \\  \hline
        Senior Manager & \cellcolor{red!40} 5000& 1111 & \cellcolor{red!40} Stark Industries\\ \hline
        Senior Accountant & 8000 & 1112 & PiedPiper \\ \hline
        Junior Engineer & 5000 & 1113 & Hooli\\ \hline
        Senior Accountant & 10000 & \cellcolor{red!40} 4111& Hooli \\ \hline
        Junior Engineer & \cellcolor{red!40} 0& 1114 & \cellcolor{red!40} - \\ \hline
    \end{tabular}}
    \label{tab:toy1}
    \vspace{-1.5em}
\end{table}

\textbf{Feature Representation.}  
%In the absence of user-defined data constraints, such as rules, we have to resort to signals and metadata that can be extracted from a dataset automatically. 
%These signals have to be general enough to capture errors that come in various shapes and forms, such as misspellings, formatting issues, missing values, token transpositions, or semantically incorrect values.
%Finding a feature representation that covers these various data error types is challenging. 
%Theoretically, for each existing error type, we can create one data constraint encoded as a feature, such as a value is a valid email address, an English word, or constrained by a functional dependency. However, the space of possible data constraints is huge and varies significantly across different datasets and even across columns of a single dataset. Furthermore, some constraints might be hidden or only apply for very specific scenarios. 
Errors come in various shapes and forms, such as misspellings, formatting issues, missing values, token transpositions, or semantically incorrect values. In order to generalize knowledge of labeled errors, we need features that describe the context of possible data errors.

For example, consider the motivational dataset in Table~\ref{tab:toy1}. Suppose the dataset contains the five errors that are marked with red background color. Certainly, one has to include features that describe how well a value fits inside a column. For example, features that identify the valid magnitude of numbers, such as ${\textit{Salary} > 500}$, or features that describe the lexical appearance of values, such as \emph{Company} is not empty, or phone numbers should have the prefix $11$. Additionally, one has to also featurize the relationship of each cell with all the other cells in the same row and encode co-occurrences. 
Consider the errors in the first tuple. In order to identify that the \emph{Salary}~value is wrong, the model has to learn a rule similar to ${\textit{Position}.\text{contains}(\textit{"Senior")} \rightarrow \textit{Salary} > 7999}$.
Finally, we might also have syntactically correct values that are still considered to be wrong according to the ground truth. For example, the company \emph{Stark Industries} is from another cinematic universe and therefore an error. However, there is no apriori rule to capture this error. In this particular case, one could draw a parallel between the first and the last tuple where both the \emph{Salary} and \emph{Company} attributes are erroneous. So, if we can encode the classification results of the other columns as features, a classifier might be able to identify that the value in the \emph{Company}~column is more likely to be erroneous if the \emph{Salary} had been already identified as an error.

%To detect the two disguised missing values in the last row as errors, one can identify single column rules, such as ${\textit{Salary} > 500}$, or ${\textit{length(Company)} > 1}$. To identify the wrong phone number, again one could come up with a pattern constraint that makes sure that each phone number has the prefix~$11$. 
%Formulating rules to detect errors in the first tuple is more challenging. In a sense, the rule that detects that the salary value is wrong would be ${\textit{Position}.\text{contains}(\textit{"Senior")} \rightarrow \textit{Salary} > 7999}$. 
%Now, there are two problems with this rule: 
%First of all, it requires the data analyst to not only identify column dependencies but also dependencies between value tokens in different columns. Furthermore, the rule is so specific that it might not hold on a larger dataset, where other senior positions receive less salary than~$7999$. 
%An approach that automatically assigns probabilities to these aspects would relieve the user from the burden of specifying such hard constraints. 
%Finally, according to the ground truth (see Definition~\ref{def:error_def}), the \emph{Company} \emph{Stark Industries} is from another cinematic universe and therefore an error. However, there is no apriori rule to capture this error.

Therefore, the first challenge is to present a general feature representation that captures a large number of error types and can be extracted automatically.

\textbf{Labeling Effort.}
Once we have a feature set that can represent each data cell of a dataset, the next challenge is to model the classification task in a way that is efficient with regard to the required number of labels.
If we employ a single classifier for the error detection task, we would have to train a multi-label classification model~\cite{tsoumakas2007multi}. Because not all cells of a tuple are correct or erroneous at the same time, such a model requires a separate label for each attribute value of a tuple.  
% that's why we use one classifier per column
Therefore, we propose to use one classifier per column, requiring a separate training set per column~\cite{tsoumakas2007multi}. Each classifier can use features from the entire dataset for classifying the values of its corresponding column.
% Active learning to sample
We can then apply active learning, which has been shown to be effective in classification tasks with imbalanced classes~\cite{ertekin2007learning}.
Using one classifier per column additionally provides us with an opportunity to share insights on classification results across classifiers.  This way, the models can detect error correlations across columns faster.
However, considering that errors in different columns are differently hard to detect, we would need a different amount of labels for each column. This problem becomes more critical as the ratio of erroneous values to correct values in a dataset is typically imbalanced, i.e., the class of erroneous values is much smaller than the class of correct values, leading to higher diffusion of data errors inside a dataset. Therefore, we need an approach that judiciously selects the most promising columns from the dataset for labeling.

To address the aforementioned challenges, we propose the cell-wise error detection system \system{} that classifies data errors with the help of the user, who progressively labels dataset cells as erroneous or correct. 
In particular, we make the following contributions:

\begin{itemize}
\item 
\begin{sloppypar}
To reduce the user labeling effort, we present a new two-dimensional multi-classifier active learning policy~(Section~\ref{sec:AL}) to progressively choose the appropriate cells for labeling. This active learning strategy advances standard active learning by not only finding the right cells within one column (one dimensional) but also across all columns (two dimensional). %We leverage the user-provided labels to automatically optimize the hyperparameters of the classification models using grid search and cross-validation.
\end{sloppypar}

\item \begin{sloppypar}To cover the wide range of error types, our new example-driven error detection approach \system{} leverages a holistic set of features~(Section~\ref{sec:features}) based on metadata, a character-level language model, value correlations, and error correlations. Furthermore, we leverage column-wise feature concatenation to exploit inter-column feature relationships.
\end{sloppypar}

\item We conduct extensive experiments~(Section~\ref{sec:evaluation}) showing that our method surpasses state-of-the-art accuracy while keeping the user effort low. Furthermore, we report how different feature sets, classification models, and labeling strategies influence the performance of our method. 
%We provide the implementation of \system{} and the datasets in our repository: \url{https://github.com/BigDaMa/ExampleDrivenErrorDetection}.
\end{itemize}

\begin{figure}
	\centering
	\includegraphics[width=\columnwidth]{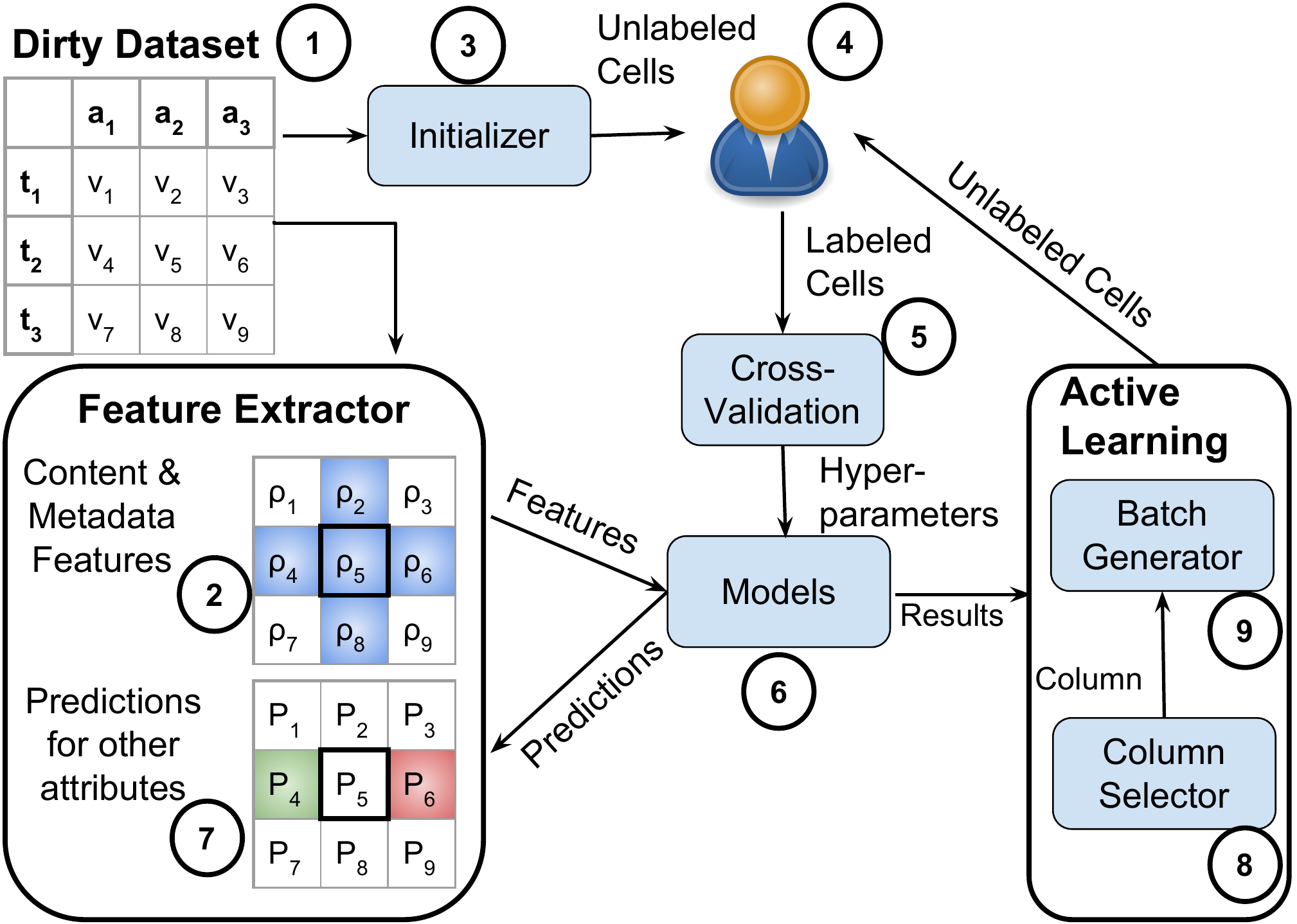}
	\caption{The \system{} workflow.}
	\label{figure:workflow}
	\vspace{-1.5em}
\end{figure}

\section{Problem Statement}
\label{sec:problem}

We address the problem of error detection in relational datasets.
We denote $D = \{t_{1}, t_{2}, ..., t_{N}\}$ as a dataset of size~$N$, where each $t_{i}$ represents a tuple. 
Let~$A = \{a_{1}, a_{2}, ..., a_{M}\}$ be the schema with attributes~$a_{j}$. Then, ${D[i, j]}$ represents the cell value of the attribute~$a_{j}$ in the tuple~$t_{i}$.
We denote $D_{C}$ as the cleaned version of $D$ representing the ground truth. 

\begin{definition}
\label{def:error_def}
Similar to existing literature~\cite{rekatsinas2017holoclean,abedjan2016detecting}, we define an error as any cell value~${D[i, j]}$ that deviates from its ground truth value~${D_{C}[i, j]}$.
\end{definition}

The goal of our method is to detect errors in a relational dataset~$D$ by letting the user progressively label attribute values as erroneous or correct.
Therefore, the first problem is to identify the right numerical feature representation~$\rho$, which effectively describes the content of each cell~${D[i, j]}$.
The second problem is to develop a sampling strategy that chooses the most promising training set to maximizes the error detection effectiveness, i.e., $F_{1}$-score. 

The $F_{1}$-score is defined as $F_1 = 2 \times (P \times R) /(P + R)$, where the precision~(P) is the fraction of cells that are correctly detected as errors and the recall~(R) is the fraction of the actual errors that are discovered.

\begin{comment}
\mohammad{remove the rest.} 

In this paper, we address both challenges with our approach that leverages a comprehensive set of features that captures syntactic and semantic information about each data cell and employs a novel active learning strategy. Our approach does neither require the user to provide dataset-specific rules or patterns nor to manually tune model-specific hyperparameters.
\end{comment}

\section{Two-Dimensional Active Learning}
\label{sec:AL}

\begin{sloppypar}

Before discussing our active learning strategy, we present the overall workflow of \system{} in Figure~\ref{figure:workflow}. 
\system{} takes a dirty dataset~$D$ as input~\ding{182}. 
The \emph{Feature Extractor} generates features~$\rho$ for each data cell of the dataset~\ding{183} as described in Section~\ref{sec:features}. 
Then, the user receives an initial sample of cells from all columns, which should be labeled as erroneous or correct~\ding{184}.
Leveraging the user-provided labels~\ding{185}, \system{} trains one classifier for each column.
\system{} leverages the user-provided labels to automatically optimize the hyperparameters of the classification models using grid search and cross-validation~\ding{186}.

Afterward, \system{} applies this model to all data cells of the corresponding column and estimates the probability of a cell to be erroneous~\ding{187}. 
In step~\ding{188}, \system{} leverages these predictions~$P$ to augment the feature vector and enables knowledge sharing across models, as we discuss in Section~\ref{sec:error_correlation_features}. 
Once the models for all columns are initialized~\ding{187}, the actual active learning process starts.

Our two-dimensional active learning policy is implemented via the \emph{Column Selector} component and the \emph{Batch Generator} component. 
As we train one classifier per column, the \emph{Column Selector} has to choose the column that should be labeled next~\ding{189}. In Section~\ref{sec:order}, we describe how the \emph{Column Selector} leverages the results of the models to make this decision. 
Then, the \emph{Batch Generator} selects the most promising cells for the given column~\ding{190} as detailed in Section~\ref{sec:uncertaintysampling}. 
For each data cell in the batch, its corresponding tuple is presented to the user for manual verification~\ding{185}. 
After the batch of cells is labeled, the new labels are added to the training set of the corresponding column, model hyperparameters are optimized~\ding{186}, and the classifier is retrained on the new data~\ding{187}.
From this point on, the process continues and repeats the steps from~\ding{185} to~\ding{190} in a loop.
\end{sloppypar}
During the entire active learning process, \system{} provides the user with a summary of the current and previous cross-validation performances, the certainty distributions, and the most decisive features per column. 
The active learning loop continues as long as the user is willing to provide additional labels. Then, the system applies the latest classification models for each column, marks the errors, and returns the result to the user. 

Next, we explain the design of our two-dimensional active learning strategy.

\subsection{Column Selection}
\label{sec:order}

As we train one classifier per column, we need a strategy for the \emph{Column Selector} to decide the next column for user labeling.
A naive approach for column selection is the random column selection strategy~(RA). This strategy randomly chooses one column and lets the user label one batch of cells for this selected column. A more orderly approach is the selection in a round-robin fashion~(RR). So, we assign the batches to the user for all columns in equal portions and in a circular order.

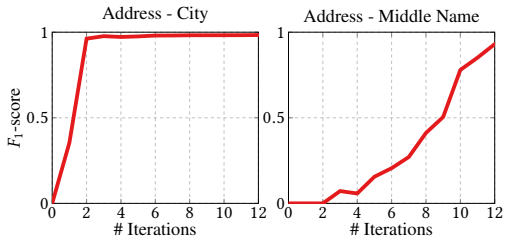
\begin{figure}[t!]
\centering
\begin{center}
\trimbox{0cm -0.1cm 0cm 0cm}{
\begin{tikzpicture}[scale=0.4, font=\Huge]
\begin{groupplot}[
    group style={
        group name=my plots,
        group size=2 by 1,
        xlabels at=edge bottom,
        ylabels at=edge left,
        horizontal sep=1cm,
        vertical sep=0cm
        },
    ytick={0.0,0.5,1.0},
    legend style={at={(0.5,-1.5)},anchor=north, legend columns=3},
    xmajorgrids=true,
    ymajorgrids=true,
    cycle list name=mycolorlist,
    grid style=dashed,
    xmin=0, xmax=12,
    ymin=0.0, ymax=1.0,
    xlabel={\# Iterations}
    ]
\nextgroupplot[title={Address - City},
ylabel={$F_1$-score}]

\addplot+[mark=none,line width=3.6pt] coordinates{(0,0.0)(1,0.35408981)(2,0.9622485)(3,0.97680797)(4,0.97212821)(5,0.97531203)(6,0.98038788)(7,0.98072717)(8,0.98206192)(9,0.98208923)(10,0.98231248)(11,0.98263704)(12,0.9833397)(13,0.98063552)(14,0.98372765)(15,0.98393509)(16,0.98407427)(17,0.98421948)(18,0.98441969)(19,0.98440883)(20,0.98455714)(21,0.98461287)(22,0.98458522)(23,0.98463969)(24,0.98462628)(25,0.98468116)(26,0.98469499)(27,0.98469499)(28,0.98469499)(29,0.98469499)(30,0.98469499)(31,0.98469499)(32,0.98469499)(33,0.98469499)(34,0.98469499)(35,0.98469499)(36,0.98469499)(37,0.98469499)(38,0.98469499)(39,0.98469499)(40,0.98469499)};

\nextgroupplot[title={Address - Middle Name}]

\addplot+[mark=none,line width=3.6pt] coordinates{(0,0.0)(1,0.0)(2,0.0)(3,0.07179487)(4,0.05780347)(5,0.15555556)(6,0.20540541)(7,0.27083333)(8,0.41148325)(9,0.5045045)(10,0.77941176)(11,0.85121107)(12,0.93037975)(13,0.93929712)(14,0.93929712)(15,0.94603175)(16,0.99697885)(17,1.0)(18,1.0)(19,1.0)(20,1.0)(21,1.0)(22,1.0)(23,1.0)(24,1.0)(25,1.0)(26,1.0)(27,1.0)(28,1.0)(29,1.0)(30,1.0)(31,1.0)(32,1.0)(33,1.0)(34,1.0)(35,1.0)(36,1.0)(37,1.0)(38,1.0)(39,1.0)(40,1.0)};
                
\end{groupplot}
\end{tikzpicture}
}

\end{center}
\caption{Example of different model convergence behavior.} 
\label{fig:column_progress}
\vspace{-1.5em}
\end{figure}

Both of the presented naive strategies do not consider that, for some columns, the model might detect errors easily and converge quickly, whereas for other columns, there might be more complex, diverse errors that cause the model to converge slower. Slower convergence also means that the user needs to label more cells to achieve the same performance. Figure~\ref{fig:column_progress} illustrates one example of significant differences between the convergence behavior of two columns. In this example, the optimal column selection strategy would choose to label the column \emph{City} only for two iterations because after that the classifier is already relatively accurate in classifying the \emph{City} values. Then, the strategy would continue to ask the user for labels for the column \emph{Middle Name}. Picking the best column not only accelerates the convergence of the classifiers but also increases the quality of some of our error correlation features as described in Section~\ref{sec:features}.

We studied three strategies to choose the next best column.

\textbf{Min Certainty (MC).} The classification model returns a probability score for each prediction, i.e., the certainty. High certainty correlates with model convergence~\cite{zhu2007active}. Thus, we calculate the average certainty for all cells in a column and choose the column with the lowest average certainty. 

\textbf{Max Error (ME).} In each iteration, we apply cross-validation on the current labeled training set for the corresponding column. The cross-validation scores are an estimate of the overall performance on the whole column data and thereby also correlate with convergence~\cite{zhu2007active}. We calculate the average of all cross-validation $F_1$-scores per column and choose the column with the minimum average cross-validation $F_1$-score.

\textbf{Max Prediction Change (MPC).} Prediction change is the fraction of predictions that changed compared to the previous iteration. Prediction change negatively correlates with active learning convergence~\cite{bloodgood2009method}. We choose the column with the highest number of prediction changes.

These column selection strategies take the convergence of the models into account to optimize the global $F_1$-score for the corresponding dataset. To gather all these metrics, we apply one round of round-robin first. Then, we can use one of these metrics to identify the next best column.

For the \emph{Column Selector}, we choose Min Certainty because it is most stable against local optima as reported in Section~\ref{subsec:orderstrategy}. For instance, if we use the Max Error strategy, we might achieve 100\%~$F_{1}$-score on the labeled training set, but still perform poorly on the overall dataset. Also, in the case of the Max Prediction Change strategy, a minor or no prediction change does not necessarily indicate that there is no learning progress. 
%The experiments in Section~\ref{subsec:orderstrategy} show that the Min Certainty strategy is effective across all datasets.

% \subsection{Stopping Criteria}
% \label{sec:stopAL}

% The users can decide on their own when to stop the active learning procedure. To support the users in their decision and to provide them with a good understanding of the current state of the dataset, we present the following statistics \mohammad{remove:} in the \emph{Status Report} at every stage of the procedure:

% \begin{itemize}
%   \item The $F_{1}$-score, precision, and recall on all labeled cells per column.
%   \item The predictions for the top $k$ least certain and most certain dataset cells per column. These insights help the user to estimate the precision of the error detection classifiers. Additionally, the user can also browse the entire error detection result.
%   \item A ranking of the most decisive features with regard to information gain for each classification model to understand better whether the underlying model works correctly.
%   \item The certainty distribution of each model for each column.
%   \item The prediction change of all previous iterations for each column. For instance, if there was no or only little prediction change over multiple iterations, the model is probably converged and further labeling is not necessary.
% \end{itemize}

% \mohammad{remove:}Based on all this information, the user has an overview of the current state of the dataset and can assess when to stop.

\subsection{Distinct Batch Sampling}
\label{sec:uncertaintysampling}
Once the next column has been selected by the Column Selector component, the next step is to select actual values within a column. 
For the data cell values within a column, we apply standard active learning techniques. In particular, we apply the query-by-committee algorithm~\cite{settles2010active} because it is a good fit for XGBoost~\cite{chen2016xgboost} that has the fastest convergence in comparison to other models, as experiments showed in Section~\ref{sec:model_selection}. 
Furthermore, we apply batch active learning~\cite{settles2010active} to reduce the runtime. 
In order to keep the sampled batch diverse, we choose cells with distinct values if possible.

\iffalse
Ideally, it makes sense to ask the user to label data cells that the model is least certain about to be erroneous. \mohammad{replace with: The basic intuition of active learning states that ask user to label those data cells that the model is least certain about their target class~[an active learning paper].}
For instance, it might not help to add another example of an error~\emph{X} if the classification model has already seen and learned the concept of this error.
To follow this approach, we select one batch of samples for one column by applying standard active learning on all values of the corresponding column. 
Specifically, we apply uncertainty sampling~\cite{settles2010active} or in the case of ensemble models, such as XGBoost~\cite{chen2016xgboost}, the query-by-committee algorithm~\cite{settles2010active}\za{Are these equivalent? Sounds like two different approaches} \felix{they are two approaches}. 
In order to keep the sampled batch diverse, we choose cells with distinct values if possible.
\fi

\subsection{Initialization}
\label{sec:init}
In order to be able to train a classifier, we need positive and negative examples for erroneous cells. As a minimum, we should start active learning with two erroneous and two correct cells per column. 
Instead of browsing through the whole dataset and identifying these cells to start up classification, we run an outlier detection method to pre-filter the possible cells. As a parameter-free variant, we employ a frequency-based ranker that ranks the values of each column by their frequency. Hence, the initial set of cells contain rare and frequent values as potential erroneous and clean training examples.

\section{Feature Representation}
\label{sec:features}
%various data types
The correct feature representation~$\rho$ is critical to identify both syntactic and semantic errors. Syntactic data errors are all values that are not in the valid domain or format of the corresponding column. For example, \emph{"Londonnn"} is not in the value domain of \emph{City}. All errors that do not violate the syntax are called semantic errors. \emph{London} as the capital of \emph{United States} is a semantic error, because the value \emph{London} is syntactically correct as a capital city but wrong in the given relationship.

%outlook
In order to detect both syntactic and semantic errors, we have to model the context of both the patterns inside the specific column and the relationships among multiple columns.
Our feature representation is a combination of existing well-known metadata features, fine-granular text features, and novel multi-column features. 
Before we describe the novel multi-column features, we review the set of single-column features that have been used in prior work.

\subsection{Single-Column Features}
We discuss two approaches to featurize a single column.
First, we present a text feature representation that can be used to detect syntactic errors. 
Then, we discuss well-known metadata features~\cite{visengeriyeva2018metadata,pit2016outlier,krishnan2017boostclean,SchelterLSCBG18} that provide additional information to extend the text feature representation.

\subsubsection{Text Features}
\label{sec:text_features}
\label{subsec:bagofcharacters}
A common way to represent text is the bag-of-words representation~\cite{harris1954distributional,krishnan2016activeclean}. Relational data does not only consist of words, but also of numbers, dates, and many more diverse types of data. Furthermore, some data errors can only be discovered at a finer granularity.
%example
For instance, the salary value \emph{"1200\$"} requires the dollar sign to be syntactically correct. 
Thus, the value \emph{"1200"} would be erroneous because it does not contain a dollar sign and violates the required syntax. A word-level representation would only learn this error for \emph{"1200"} and would not be able to generalize it to other salary values.
%If the user labeled the value \emph{"1200"} as erroneous and we train a model on the word-level representation for this value, the model might classify \emph{"2000"} as correct because we do not have any information about the word \emph{"2000"} yet. 
Using a character-level representation, the model will learn that the absence of the character \emph{'\$'} correlates with erroneous syntax. Thus, another salary value \emph{"2000"} would be classified as erroneous because the value also does not contain a dollar sign.
Therefore, instead of a word-level language model as used in prior work~\cite{krishnan2016activeclean}, we use the more fine-granular character-level language model~$\rho_{\text{n-grams}}$.

\begin{sloppypar}
\label{subsec:bagofcharacters}
The \textbf{bag-of-n-grams model} is a character-based n-gram language model~\cite{cavnar1994n}. Here, the \emph{Feature Extractor} counts all character sequences of length~$n$ occurring in a given column of the dataset. In the case of a unigram model~($n=1$), the \emph{Feature Extractor} counts the occurrences of single characters, such as \mbox{'A'}, \mbox{'a'}, \mbox{'1'}, \mbox{'.'}, and~\mbox{' '}, for each cell~${D[i,j]}$ of a given column ${D[:,j]}$. The \emph{Feature Extractor} applies the common approach of the TF-IDF score to normalize the n-gram counts~\cite{sparck1972statistical}. Formally, we define the bag-of-n-grams feature vector as
\end{sloppypar}

\begin{equation} \label{equation:feature_ngrams}
\resizebox{\hsize}{!}{
	$\rho_{\text{n-grams}}(D[i,j],n) = [ \tf(\omega, D[i,j]) \times \idf(\omega, D[:,j]) | \omega \in \Omega_{n}(D[:,j]) ],$
}
\end{equation}
\begin{sloppypar}
where $\Omega_{n}(D[:,j])$ is the set of all n-grams~$\omega$ of length~$n$ that exist in column~${D[:,j]}$. 
The term frequency~$\tf$ measures the number of times that a specific n-gram~$\omega$ occurs in the cell~${D[i,j]}$ and the inverse document frequency~$\idf$ measures whether a specific n-gram~$\omega$ is common or rare across all cells in the column~${D[:,j]}$.

%Therefore, the feature representation~$\rho_{\text{n-grams}}$ for each data cell~$D[i,j]$ is the list of the TF-IDF scores of all n-grams~$\omega$ in the corresponding column~$D[:,j]$.
%Choosing $n$ is not trivial. The unigram text representation~(${n=1}$) does not provide any order information, such as which character frequently succeeds or precedes certain other characters. Increasing the \mbox{n-gram} length to ${n>1}$ provides order information within the \mbox{n-gram} model. However, increasing the \mbox{n-gram} length also results in a significant expansion of the number of features that causes longer model training and overall runtime. 
%Therefore, we try to find a trade-off that works well for most of the datasets as discussed in Section~\ref{sec:feature_selection_experiment}. 
\end{sloppypar}

\subsubsection{Metadata Features}
\label{sec:explain_metadata_features}
Metadata can be used to complement the text features with domain-independent features that can be used to describe errors. Previous studies have shown the applicability of various metadata and statistics for data quality assessment~\cite{SchelterLSCBG18}.

\begin{sloppypar}
We leverage previously proposed metadata features: the value occurrence count~$\rho_{\text{occurrence}}$~\cite{visengeriyeva2018metadata}, the string length of a value~$\rho_{\text{string\_length}}$~\cite{pit2016outlier}, the data type of a value~$\rho_{\text{data type}}$~\cite{visengeriyeva2018metadata}, and the numeric representation of a value~$\rho_{\text{number}}$~\cite{krishnan2017boostclean}.
We combine these features in a metadata feature vector of a data cell as follows:
\end{sloppypar}

\begin{footnotesize}
	\begin{equation} \label{equation:feature_meta}
	\begin{split}
	\rho_{\text{meta}}(D[i,j]) = \Big[\rho_{\text{occurrence}}(D[i,j]), \rho_{\text{string\_length}}(D[i,j]), \\ \rho_{\text{data type}}(D[i,j]), \rho_{\text{number}}(D[i,j])\Big].
	\end{split}
	\end{equation}
\end{footnotesize}

%Of course, this small set of features is not exhausting the space of all possible features, but it is general enough to capture fundamental characteristics of most datasets. 
%Additionally, the user can add more dataset-specific features. 

\subsection{Feature Aggregation and Multi-Column Features}
Correlations and value relationships across columns can support the error detection process. Since we have to train one classifier per column, there are several challenges and opportunities to capture inter-column relationships.
We introduce three approaches to embed signals from other columns as features for a column classifier at hand. 
First, we adopt word embedding techniques to encode the semantic relationship of dataset values across columns.
Second, we use a concatenation approach to model the relationship of labeled data cells in one column with individual features of other columns.
%\felix{Second, we use a concatenation approach to allow the column classifier to leverage features across columns that correlate with the user-provided labels at hand.}
Third, \system{} leverages prediction results of other classifiers to obtain error correlations within a tuple.

\begin{figure}[t!]
	\centering
	%\begin{table}[h!]
	%\centering
	%\caption{Example for an error that can be detected only by exploiting inter-column correlations.}
	%\label{tab:columnword2vec}
	\begin{tabular}{|l|r|} \hline
		Position & Salary \\ \hline
		senior manager & \cellcolor{red!50} 5,000 \\ \hline
		senior manager & 10,000 \\ \hline
		senior manager & 10,000 \\ \hline
	\end{tabular}
%\end{table}
\hfill
	%\begin{figure}[h]
\begin{tikzpicture}[baseline=(current bounding box.center), scale=0.6, every node/.style={transform shape}]
    \tkzInit[xmax=5,ymax=4,xmin=0,ymin=0]
    \tkzDrawX[label={}]
    \tkzDrawY[label={}]
    \draw[red,thick,dashed] (3.5,3) circle (1.5);
    \coordinate (a) at (1.5,1);
    \coordinate (b) at (3.5,3);
    \coordinate (c) at (4.5,4);
    \node[label={left:\Large 5,000}] at (a)  {};
    \node[label={left:\Large senior manager}] at (b) {};
    \node[label={left:\Large 10,000}] at (c) {};
    \draw (a) -- (c);
    \foreach \Nombre in {a,b,c}
      \node[circle,inner sep=1.5pt,fill=black] at (\Nombre) {};
\end{tikzpicture}
%\end{figure}
	\caption{An example to show how the cell value embedding can help to detect semantic errors leveraging cell value correlations.}
	\label{figure:exampleword2vecembedding}
	\vspace{-1.5em}
\end{figure}
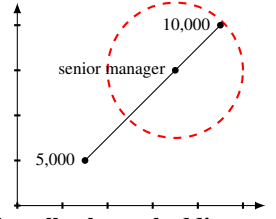

\subsubsection{Cell Value Correlation Features}
%Column-wise feature concatenation requires labels to uncover correlations and to exploit them for error detection. 
%If a correlation that would enable the classifier to identify an error is not detectable using the labeled training set alone, column-wise feature concatenation might fail because the classifier can only access the labeled training set. 
%\mohammad{We can also leverage the nature structure of relational data to learn the cell value correlations.} 

Word embeddings are a novel technique for identifying the semantic similarity of values~\cite{mikolov2013distributed}.
One can apply the same technique to identify correlations among values inside a dataset.
To model all cell value correlations of the whole dataset, we use the \emph{Word2vec}~\cite{mikolov2013distributed} features. We consider each tuple as a "document" and each cell value~${D[i,j]}$ as a "word"~\cite{krishnan2017boostclean}. Then, we learn a word embedding~$\rho_{\text{w2v}}$ that maps each cell value~${D[i,j]}$ into a multidimensional space $\rho_{\text{w2v}}(D[i,j])$ where co-occurring cell values are close to each other.

%Figure~\ref{figure:exampleword2vecembedding} illustrates an example of this approach. The colored cells represent user labels. In this case, using the labeled training set alone, we lose the information that the combination of the position \emph{"senior manager"} and the salary \emph{"10{,}000"} is frequent. This information might help the classifier to better differentiate against outlying combinations, such as the salary \emph{"5{,}000"} and the position \emph{"senior manager"}. 
As shown in Figure~\ref{figure:exampleword2vecembedding}, the classifier could consider all \emph{Salary} values within the vicinity of the embedding of the \emph{Position} value as correct, i.e., \emph{"10{,}000"} is a correct \emph{Salary} for \emph{"senior manager"}.
%For instance, \emph{"10{,}000"} co-occurs two times with \emph{"senior manager"} while \emph{"5{,}000"} co-occurs with \emph{"senior manager"} only once. Therefore, \emph{"10{,}000"} is also closer in the embedding to \emph{"senior manager"} than \emph{"5{,}000"}. Furthermore, \emph{"10{,}000"} and \emph{5{,}000} do not co-occur and therefore are very far apart in the embedding.

%Finally
%While the \emph{Word2vec} approach models correlations among cell values, we still need column-wise feature concatenation because it can be used to exploit correlations among metadata and character-level text features. \mohammad{is this explanation necessary?} Do you think it is clear or does it confuse you? i think it is clear but even if it is not, i do not think reviewer would ask this question.ok
%what happens for out-of-vocabulary words
%Finally, if the user wants to apply these features for new unseen dataset that contains new out-of vocabulary words, we can retrain the Word2vec model on all new data and then retrain our classification model. \za{Do we need this statement? It is surprising to me}

%\mohammad{remove:} However, we focus on static data to this point and consider this part as future work.

\begin{figure}[t!]
	\centering
	%\begin{table}[h!]
%	\centering
%	\caption{Example for an error that can be detected only by exploiting inter-column correlations.}
%	\label{tab:columnconcat}
	\begin{tabular}{|l|r|} \hline
		Position & Salary \\ \hline
		senior accountant & \cellcolor{red!50}5,000 \\ \hline
		senior manager & 10,000 \\ \hline
		junior engineer & 4,000 \\ \hline
		senior engineer & 11,000 \\ \hline
	\end{tabular}
%\end{table}
\hfill
	\tikzset{
  treenode/.style = {shape=rectangle, rounded corners,
                     draw, align=center,
                     top color=white, bottom color=blue!20},
  root/.style     = {treenode, font=\large, bottom color=blue!20},
  correct/.style  = {treenode, font=\large, bottom color=green!20},
  error/.style    = {treenode, font=\large, bottom color=red!30},
  dummy/.style    = {circle,draw},
  level 1/.style  = {level distance=9em},
  level 2/.style  = {level distance=11em}
}
\begin{tikzpicture}
  [
    baseline                = (current bounding box.center),
    scale                   = 0.53,
    grow                    = right,
    sibling distance        = 4em,
    edge from parent/.style = {draw, -latex},
    every node/.style       = {transform shape},
    sloped
  ]
  \node [root] {Salary}
    child { node [correct] {correct}
      edge from parent node [below] {else} }
    child { node [dummy] {}
      child { node [correct] {correct}
        edge from parent node [below] {else} }
      child { node [error] {erroneous}
              edge from parent node [above, align=center]
                {$\text{Salary} < 10{,}000$}}
              edge from parent node [above] {'s' in Position} };
\end{tikzpicture}
	\caption{Column-wise feature concatenation: All salary values in this example have been labeled. Thus, we can use the relationship to the single-column features in Position.}
	\label{figure:columnconcat}
	\vspace{-1.5em}
\end{figure}
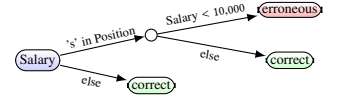

\subsubsection{Column-Wise Feature Concatenation}
\label{sec:concatenation}
\begin{sloppypar}
%why are word embedding not enough
While word embeddings can be used to identify correlations at the cell value level, they do not capture correlations among more fine-granular features. For example, they cannot represent a correlation between a character~\emph{","} in column A and the value string length of column B.
%example where word embedding would not work at all
%Furthermore, cell value correlation features cannot learn the co-occurrence of cell values in the absence of data redundancy.
The dataset in Figure~\ref{figure:columnconcat} contains only distinct value pairs of \emph{Salary} and \emph{Position}. Therefore, there is no cell value co-occurrence.
%so how could column concatenation help here
In order to learn that the salary of \emph{"senior"}~positions is significantly higher than that of \emph{"junior"}~positions, the classification model has to have access to information that concerns both \emph{Salary} and \emph{Position}.
By concatenating the single-column features of all columns, we enable the classifier to model these fine-granular inter-column relationships.

%example for a model that uses this feature concatenation
In Figure~\ref{figure:columnconcat}, we show a decision tree that was trained based on the concatenated features of both columns~\emph{Salary} and \emph{Position}. This model successfully identified the rule that if the \emph{Position}~value contains an \emph{"s"} for \emph{"senior"} and the \emph{Salary}~value is smaller than \emph{"10{,}000"} then the \emph{Salary}~value is erroneous.
%how do we do this in general?
Since we do not know upfront which inter-column relationships exist in the corresponding dataset, we concatenate features from all columns and let the classifier decide which of the features are useful.

%While some of these relationships can also be modelled by cell value correlation features introduced before, column-wise feature concatenation can also express correlations between metadata and character-level text features, which are not captured by word embeddings.
Formally, we concatenate the single-column features of all $M$~columns as follows:
\end{sloppypar}

\begin{scriptsize}
\begin{equation} \label{equation:feature_concat}
\rho_{\text{concat}}(D[i,j], \rho) = \Big[\rho(D[i,1]), \dots, \rho(D[i,M])\Big].
\end{equation}
\end{scriptsize}

%This concatenation enables the classifier to detect errors that can be exposed via value dependencies across multiple columns. 
%\mohammad{remove:} So, the machine learning model has not only access to the content of the cell in question, but also to the content of all other cells of the corresponding tuple.
%\mohammad{I liked this section.} \felix{I am happy}

\iffalse
\begin{figure}[t!]
	\centering
	\input{../tables/example_error_correlation}\hfill
	\caption{An example that shows encoding issues across attributes of instances.}
	\label{figure:error_correlation_example}
\end{figure}
\fi

\subsubsection{Error Correlation Features}
\label{sec:error_correlation_features}
%Daniel Defoe:
Misfortune seldom comes alone --- the same often applies to errors. Once, a classifier detects an error in one of the attributes of a tuple, it is more likely that there is an error in another attribute of the same tuple as well.
In particular, there is error co-occurrence when certain tuples come from an unreliable source. For instance, in Table~\ref{tab:toy1}, we illustrate a potential error correlation.
Whenever the \emph{Salary} value is erroneous, the \emph{Company} value is erroneous as well.

So far, each classification model can access only the labels of its corresponding column $j$ and therefore does not know about errors in the other columns $k$.
We want to incorporate knowledge about errors in other columns. 
Let $\phi_{k}$ be the last trained model on the data column~$k$. 
Further, let $P_{\phi_{k}}(D[i, k] = erroneous)$ be the estimated error probability of the data cell~${D[i, k]}$ based on the model $\phi_{k}$.
Therefore, for our current column $j$, we concatenate the estimated error probabilities of all other columns~$k \neq j$. Formally,

\begin{scriptsize}
\begin{equation} \label{equation:feature_errorcorrelation}
\begin{split}
\rho_{\text{error}}(D[i,j]) = \Big[P_{\phi_{1}}(D[i, 1] = \text{erroneous}), \dots , P_{\phi_{j-1}}(D[i, j-1] = \text{erroneous}), \\ 
P_{\phi_{j+1}}(D[i, j+1] = \text{erroneous}) , \dots , P_{\phi_{M}}(D[i, M] = \text{erroneous})\Big].
\end{split}
\end{equation}
\end{scriptsize}

In case of very small error fractions, it is harder to find examples for erroneous values. Our experiments show that particularly in such cases, the error correlation features improve the prediction quality of the classifiers. We add these features to the feature matrix in step~\ding{188} of Figure~\ref{figure:workflow}.

\subsection{Feature Vector at a Glance} 
\label{sec:final_features}

The \emph{Feature Extractor} concatenates all the features that can be obtained per column, i.e., the n-gram features, metadata features, Word2vec features, together with the error correlation features that are extracted after a prediction step. Formally, we can define the final feature vector per data cell~${D[i,j]}$ as follows:

\begin{scriptsize}
\begin{equation} \label{equation:feature_all}
\begin{split}
\rho_{\text{\system{}}}(D[i,j]) = \bigg[\rho_{\text{concat}}(D[i,j], \rho_{\text{n-grams}}), \rho_{\text{concat}}(D[i,j], \rho_{\text{meta}}), \\
\rho_{\text{concat}}(D[i,j], \rho_{\text{w2v}}), \rho_{\text{error}}(D[i,j])\bigg].
\end{split}
\end{equation}
\end{scriptsize}

\subsection{Explainability}
\label{sec:stopAL}

\begin{sloppypar}
One of the reasons that data practitioners still resort to rule-based approaches is that rules are explainable. 
In order to explain classification results, we train simple interpretable models, such as a decision tree. By training this model on our fine-granular features, \system{} is able to provide the decision path as explanation to the user. For instance, the explanation for the erroneous value \emph{Stark Industries} in Table~\ref{tab:toy1} would be the following decision path: IF ${P(\text{Salary("5000")} = \text{erroneous}) > 0.8}$ THEN \emph{Stark Industries} is erroneous. 
If the user detects a mistake of the classification model, the user can also prevent the model from using specific features to improve and accelerate model learning. 
\end{sloppypar}

\section{Experiments}
\label{sec:evaluation}

\begin{figure*}[ht!]
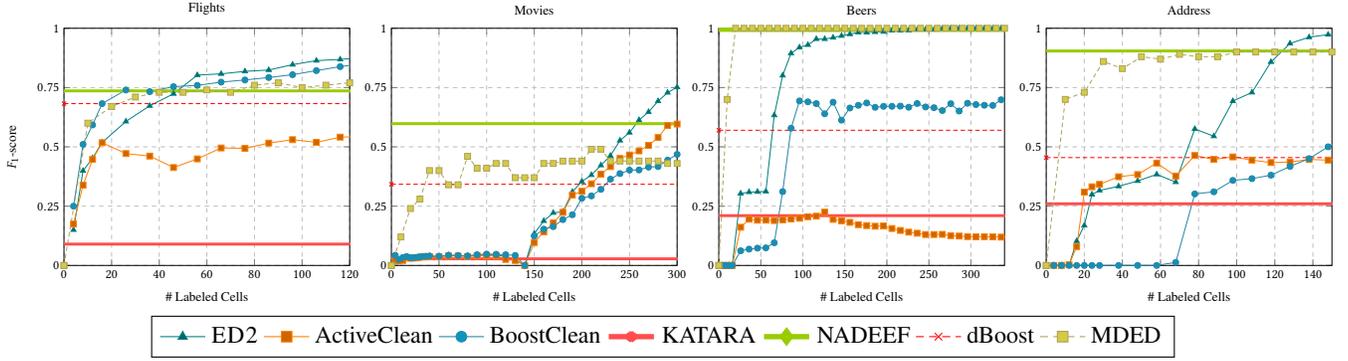

	\centering
	\begin{center}
\trimbox{0cm -0.1cm 0cm 0cm}{
\resizebox{\textwidth}{!}{
\begin{tikzpicture}
\begin{groupplot}[
    group style={
        group name=my plots,
        group size=4 by 1,
        xlabels at=edge bottom,
        ylabels at=edge left,
        horizontal sep=1cm,
        vertical sep=1.5cm,
        },
    ytick={0.0,0.25,0.5,0.75,1.0},
    legend style={at={(0.5,-1.5)},anchor=north, legend columns=-1},
    xmajorgrids=true,
    ymajorgrids=true,
    cycle list name=exotic,
    grid style=dashed,
    xlabel={\# Labeled Cells},
    xmin=0.0, ymin=0.0, ymax=1.0
    ]

\input{charts/labels_all/labels_flights}
\input{charts/labels_all/labels_movies}
\input{charts/labels_all/labels_beers}
\input{charts/labels_all/labels_address}
                
\end{groupplot}
\end{tikzpicture}
}
}
\ref{testLegendlabels}
\end{center}
	\caption{$F_1$-score and labeling effort of different error detection methods for different datasets.}
	\label{figure:methods_comparison_user_effort}
	%\vspace{2mm}
\end{figure*}

\newcommand{\ra}[1]{\renewcommand{\arraystretch}{#1}}
\begin{table*}\centering
\caption{Best $F_1$-score results of each method in Figure~\ref{figure:methods_comparison_user_effort} and their corresponding precision (P) and recall (R).}
\ra{1.3}
\resizebox{\textwidth}{!}{
\begin{tabular}{@{}lrrrcrrrcrrrcrrr@{}}\toprule
& \multicolumn{3}{c}{Flights} & \phantom{abc}& \multicolumn{3}{c}{Movies} &
\phantom{abc} & \multicolumn{3}{c}{Beers} &
\phantom{abc} & \multicolumn{3}{c}{Address}
\\
\cmidrule{2-4} \cmidrule{6-8} \cmidrule{10-12} \cmidrule{14-16} 
& P & R & $\text{F}_1$ && 
P & R & $\text{F}_1$ && 
P & R & $\text{F}_1$ && 
P & R & $\text{F}_1$
\\ \midrule
\system & 
    0.84 $\pm$ 0.03 & 0.90 $\pm$ 0.02 & \textbf{0.87} $\pm$ 0.01 && %flights 116
	%0.73 & 0.76 & 0.74 & %citations  112
	0.89 $\pm$ 0.16  & 0.67 $\pm$ 0.13 & \textbf{0.75}  $\pm$ 0.12&& %movies 300
	%0.93 $\pm$ 0.02 & 0.69 $\pm$ 0.08 & 0.79 $\pm$ 0.06 && % restaurants 499
	1.00 $\pm$ 0.00 & 1.00 $\pm$ 0.00 & \textbf{1.00} $\pm$ 0.00 && % beers 336 
	0.99 $\pm$ 0.01 & 0.96 $\pm$ 0.01 & \textbf{0.97} $\pm$ 0.01 % address 148
	%1.00 $\pm$ 0.01 & 0.90 $\pm$ 0.05 & \textbf{0.95} $\pm$ 0.03 
	\\ %hospital
ActiveClean & 
    0.61 $\pm$ 0.03 & 0.50 $\pm$ 0.10 & 0.54 $\pm$ 0.06 && %flights 116 # 0.605769484446,0.501016260163,0.541041580464
	%0.72 & 0.75 & 0.74 && %citations  112 0.723062923527,0.748206751055,0.73501271227
	0.73 $\pm$ 0.11 & 0.51 $\pm$ 0.07 & 0.60 $\pm$ 0.08 && %movies 300 #0.734023657596,0.506395736176,0.59608046538
	%0.79 $\pm$ 0.09 & 0.81 $\pm$ 0.02 & \textbf{0.80} $\pm$ 0.05 && % restaurants 499,0.785868242605,0.813982300885,0.797143035745
	0.43 $\pm$ 0.08 & 0.07 $\pm$ 0.03 & 0.12 $\pm$ 0.04 && % beers 336 # 0.430914281334,0.0722848360656,0.119369442751
	0.49 $\pm$ 0.05 & 0.42 $\pm$ 0.07 & 0.44 $\pm$ 0.02 % address 148 # 0.489931067146,0.417236547978,0.444133519999
	%0.09 $\pm$ 0.01 & 0.39 $\pm$ 0.02 & 0.15 $\pm$ 0.02 
	\\ %hospital
BoostClean & 
    0.83 $\pm$ 0.02 & 0.85 $\pm$ 0.03 & 0.84 $\pm$ 0.02 && %flights 116 # 0.825646075571,0.85174796748,0.838165878228
	%0.73 & 0.76 & \textbf{0.75} & %citations  112 # 0.731172922417,0.760232067511,0.745377526113
	0.60 $\pm$ 0.16 & 0.42 $\pm$ 0.10 & 0.47 $\pm$ 0.06 && %movies 300 #0.598661574254,0.424583610926,0.46829440364
	%0.90 $\pm$ 0.11 & 0.21 $\pm$ 0.07 & 0.34 $\pm$ 0.09 && % restaurants 499,679.693699574,0.896156423853,0.213982300885,0.335487112886
	0.99 $\pm$ 0.04 & 0.54 $\pm$ 0.03 & 0.70 $\pm$ 0.02 && % beers 336 # 0.985045248988,0.54262295082,0.698493835377
	0.67 $\pm$ 0.06 & 0.40 $\pm$ 0.07 & 0.50 $\pm$ 0.06  % address 148 # 0.665755896321,0.404913355538,0.500186227953
	%0.45 $\pm$ 0.17 & 0.49 $\pm$ 0.04 & 0.45 $\pm$ 0.12 
	\\
	
MDED & 
    0.81 $\pm$ 0.10 & 0.77 $\pm$ 0.14 & 0.77 $\pm$ 0.04 && %flights 116 # 0.825646075571,0.85174796748,0.838165878228
	%0.73 & 0.76 & \textbf{0.75} & %citations 
	0.68 $\pm$ 0.44 & 0.53 $\pm$ 0.14 & 0.43 $\pm$ 0.26 && %movies 300
	%0.00 & 0.00 & 0.00 && % restaurants 499
	1.00 $\pm$ 0.00 & 1.00$ \pm$ 0.00 & \textbf{1.00} $\pm$ 0.00 && % beers 336
	1.00 $\pm$ 0.00 & 0.82 $\pm$ 0.00 & 0.90 $\pm$ 0.00  % address 148 
	%0.95 $\pm$ 0.01 & 0.83 $\pm$ 0.02 & 0.88 $\pm$ 0.01 
	\\ %hospital

dBoost & 
	0.52 $\pm$ 0.00 & 1.00 $\pm$ 0.00 & 0.68 $\pm$ 0.00 && %flights
	%0.18 & 0.27 & 0.22 & %citations
	0.24 $\pm$ 0.00 & 0.63 $\pm$ 0.00 & 0.34 $\pm$ 0.00 && %movies
	%0.11 & 0.10 & 0.10 && % restaurants
	0.40 $\pm$ 0.00 & 1.00 $\pm$ 0.00 & 0.57 $\pm$ 0.00 && % beers 0.24778917580474, 0.7177254098360656, 0.3683933736523797
	0.43 $\pm$ 0.00 & 0.49 $\pm$ 0.00 & 0.45 $\pm$ 0.00 % address
	%0.90 & 0.83 & 0.86 
        \\

\iffalse
Gaussian & 
	0.37 & 1.00 & 0.54 && %flights
	%0.10 & 1.00 & 0.19 & %citations
	0.24 & 0.53 & 0.33 && %movies
	%0.01 & 0.05 & 0.02 && % restaurants
	0.25 & 0.72 & 0.37 && % beers 0.24778917580474, 0.7177254098360656, 0.3683933736523797
	0.27 & 0.64 & 0.38 && % address
	%0.83 & 0.68 & 0.75
	\\ % hospital
	
	Histogram & 
	0.50 & 0.73 & 0.59 && %flights
	%0.18 & 0.19 & 0.19 & %citations
	0.17 & 0.33 & 0.23 && %movies
	%0.11 & 0.10 & 0.10 && % restaurants
	0.40 & 1.00 & 0.57 && % beers # 0.3965867533522958, 1.0, 0.5679371544951993
	0.19 & 0.41 & 0.26 && % address
	%0.90 & 0.83 & 0.86
	\\ % hospital
	
	Mixture & 
	0.52 & 1.00 & 0.68 && %flights
	%0.18 & 0.27 & 0.22 & %citations
	0.24 & 0.63 & 0.34 && %movies
	%0.00 & 0.47 & 0.00 && % restaurants
	0.95 & 0.13 & 0.23 && % beers 0.9477611940298507, 0.13012295081967212, 0.22882882882882882
	0.43 & 0.49 & 0.45 && address
	%0.03 & 0.90 & 0.06 
	\\ % hospital
\fi
	
	KATARA & 
	0.09 $\pm$ 0.00 & 0.09 $\pm$ 0.00 & 0.09 $\pm$ 0.00 && %flights
	%0.03 & 0.10 & 0.04 & %citations
	0.02 $\pm$ 0.00 & 0.16 $\pm$ 0.00 & 0.03 $\pm$ 0.00 && %movies
	%0.00 & 0.37 & 0.00 && % restaurants
	0.13 $\pm$ 0.00 & 0.58 $\pm$ 0.00 & 0.21 $\pm$ 0.00 && % beers 0.12857791912766925, 0.5799180327868853, 0.21048716995165492
	0.18 $\pm$ 0.00 & 0.53 $\pm$ 0.00 & 0.26 $\pm$ 0.00  % address
	%0.05 & 0.25 & 0.09 
	\\ % hospital
	
	NADEEF & 
	0.79 $\pm$ 0.00 & 0.69 $\pm$ 0.00 & 0.74 $\pm$ 0.00 && %flights
	%0.70 & 0.08 & 0.15 & %citations
	1.00 $\pm$ 0.00 & 0.43 $\pm$ 0.00 & 0.60 $\pm$ 0.00 && %movies
	%0.00 & 0.00 & 0.00 && % restaurants
	1.00 $\pm$ 0.00 & 1.00 $\pm$ 0.00 & \textbf{1.00} $\pm$ 0.00 && % beers
	1.00 $\pm$ 0.00 & 0.83 $\pm$ 0.00 & 0.90 $\pm$ 0.00  % address
	%1.00 & 0.28 & 0.44 
	\\
\bottomrule
\end{tabular}
}
\label{tab:effectivity}
\vspace{-4mm}
\end{table*}

\begin{sloppypar}
We conducted several experiments to compare our approach against state-of-the-art error detection methods on multiple datasets and to benchmark different aspects of our solution. In particular, we investigate how our approach compares to state-of-the-art error detection methods in terms of effectivity and user effort. Further, we explore the influence of different feature representations, column selection strategies, and classification models on the performance of \system{}. Additional experiments with additional error detection systems and datasets can be found in our peer-reviewed paper~\cite{neutatz2019ed2}.
\end{sloppypar}

\subsection{Experimental Setup}
\label{sec:experimental_setup}
In the following, we present the datasets, competing methods, evaluation methodology, parameter configuration, and implementation details.

\subsubsection{Datasets}

We conducted our experiments on four real-world datasets that mostly have been used in prior work in data cleaning. 
Table~\ref{tab:datasets} lists these datasets along with the number of rows, number of columns, and the corresponding fraction of erroneous cells divided by all cells in the dataset. 
%As shown, the datasets vary in both shape and number of errors.

\begin{table}[H]
\small
\centering
\caption{Experimental datasets.}
\label{tab:datasets}

\begin{tabular}{@{}lrcrcr@{}}\toprule
Dataset& Columns && Rows && Errors\\ \midrule
Flights  	& 6  	&& 2,376	&& 35\%\\
%Citations	& 11& 8 	& 1,000	&  8.62\%\\
Movies		& 17 	&& 7,390	&&  1\%\\
%Restaurants	& 16&& 10 	&& 28,787	&&  0.12\%\\ 
Beers		& 10 	&& 2,410	&&  8\%\\
Address		& 12 	&& 94,306	&& 14\%\\
%Hospital	& 17&& 17 	&& 1,000	&&  3\%\\
\bottomrule 
\end{tabular}
\vspace{-1.0em}
\end{table}

\textbf{Flights.} Flights is a real-world dataset~\cite{li2012truth,rekatsinas2017holoclean}. 
Errors comprise mostly wrong departure/arrival times that violate inter-column dependencies and missing values.
\textit{We use this dataset to examine how stable our algorithm is in the presence of a large number of errors.}

\textbf{Movies.} Movies is a real-world collection of movies from Rotten Tomatoes and IMDB~\cite{magellandata}. Errors are mostly due to formatting issues. 
%For instance, IMDB reports the year range, e.g., \emph{"2005 2006"}, instead of one specific year \emph{"2005"} and the duration format is \emph{"1 hr. 25 min."} instead of \emph{"85 min"} for Rotten Tomatoes. Furthermore, identifiers and the rating differ between the two different sources. 
%\textit{We use this dataset to examine the ability of \system{} to capture heterogeneity issues of integrated datasets.}
\textit{We use this dataset to examine if \system{} can find errors in the case of a very small error fraction.}

%\textbf{Restaurants.} Restaurants is a real-world collection of restaurants from Yellow Pages and Yelp~\cite{magellandata}. Errors are formatting issues, such as missing \emph{"http://"} for websites, categorizing restaurants into \emph{"Pizza Restaurant"} versus \emph{"Pizza"}, or different address formats, e.g., \emph{"1725 E Osborn Rd"} instead of \emph{"1725 East Osborn"}. \textit{We use this dataset to examine if \system{} can find errors in the case of a very small error fraction.}

\textbf{Beers.} Leveraging web scraping, multiple types of beer have been collected~\cite{beers_dataset}. The errors are missing values, field separation issues, and formatting issues. 
\textit{We use this dataset to examine the ability of \system{} in capturing web scraping originated issues.}

\textbf{Address.} Address is an anonymized proprietary address dataset. Errors concern spelling, formatting, completeness, and field separation. \textit{We use this dataset to examine the ability of \system{} in capturing various error types in the same dataset.}

%\textbf{Hospital.} Hospital is one of the most commonly used real-world datasets to benchmark data cleaning algorithms~\cite{rekatsinas2017holoclean, chu2013holistic, dallachiesa2013nadeef}. The data originates from the US Department of Health \& Human Services. Errors are introduced synthetically by randomly replacing characters by the letter 'x'. Rekatsinas et al. provided us with both the dataset and the corresponding ground truth~\cite{rekatsinas2017holoclean}. \textit{We use this dataset to show that our method can achieve state-of-the-art accuracy on a well-established error detection benchmark.}

\subsubsection{Baselines}
We evaluate \system{} against three conventional and three novel machine learning-based competing error detection approaches.

\textbf{NADEEF}\label{nadeef_config}~\cite{dallachiesa2013nadeef} is a rule violation detection system that allows users to specify multiple types of rules to detect data errors. \textbf{Usage:}
%We run NADEEF leveraging functional dependencies and denial constraints. 
In addition to provided constraints by the data owners, we run Metanome~\cite{papenbrock2015data} to mine functional dependencies that are added to the set of constraints if they increase the overall $F_1$-score.
%All 28 constraints that were discovered for datasets in total were validated by four computer scientists. %and are described in our repository\footnote{\url{https://github.com/BigDaMa/ExampleDrivenErrorDetection}}. 
We consider any cell that participates in at least one violation to be erroneous~\cite{abedjan2016detecting}.

%Note that to use rule-based methods, the user has to be aware of existing correct constraints. For \system{}, the user neither has to understand the concept of functional dependencies nor how to detect and formulate them.  

\textbf{dBoost~\cite{pit2016outlier}} is a framework that provides various machine learning models to detect outliers. 
In contrast to traditional outlier detection methods that are built for numerical data, dBoost proposes tuple expansion to leverage rich information across data types. \textbf{Usage:}
We report the dBoost model with the highest $F_1$-score after extensive hyperparameter optimization.  

%Note that to use outlier detection methods, the users have to implement hyperparameter optimization by themselves. For \system{}, hyperparameter optimization is running without user involvement, as we explain in Section~\ref{sec:uncertaintysampling}.

\textbf{KATARA~\cite{chu2015katara}} can be used to detect semantic pattern violations using external knowledge bases. \textbf{Usage:}
We provide KATARA with access to DBpedia~\cite{lehmann2015dbpedia} that contains $2017$ relations, such as \emph{has\_ZIP} and \emph{has\_Director}. Thus, potentially it can find violations for the data domains that are present in DBpedia. 

%Note that, in contrast to KATARA, our approach does not require the user to provide external resources.

\textbf{ActiveClean~\cite{krishnan2016activeclean}} is a machine learning model training framework that allows for iterative data cleaning. It contains an adaptive error detection component based on classification. \textbf{Usage:}
We can not use the sampling approach of ActiveClean because it requires a user-specified machine learning task. 
Therefore, we apply our sampling approach while using the features proposed by ActiveClean.

\textbf{BoostClean~\cite{krishnan2017boostclean}} is an automated data cleaning system for machine learning datasets. \textbf{Usage:}
The same restrictions that apply to ActiveClean, apply to BoostClean as well. Therefore, we use our sampling approach while leveraging the features proposed by BoostClean.

\textbf{Metadata-driven error detection (MDED)~\cite{visengeriyeva2018metadata}} aggregates the results of manually configured stand-alone error detection methods and metadata to classify errors. \textbf{Usage:}
We provide this approach with the results from the traditional methods NADEEF, KATARA, and dBoost.

%Note that, instead of no configuration as for \system{}, metadata-driven error detection requires the configuration of all methods that it leverages.

Note that these methods require configuration, such as rules, hyperparameters, and external data, whereas \system{} does not.

\begin{figure*}[h!]
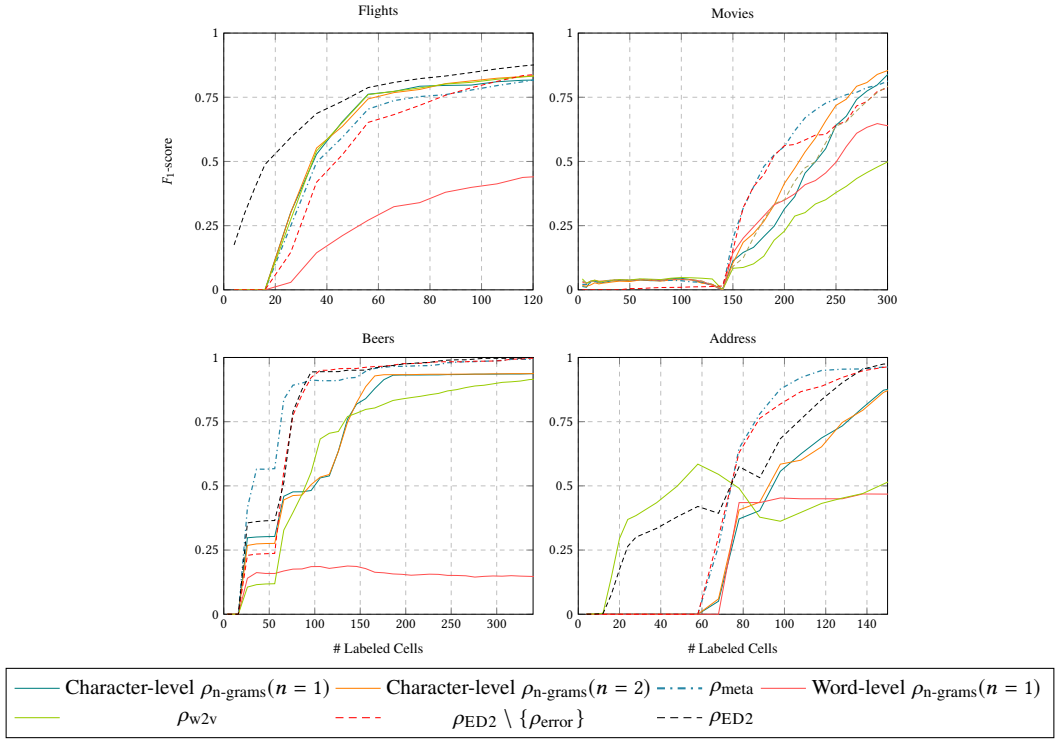

	\centering
	\begin{center}
\trimbox{0cm -0.1cm 0cm 0cm}{
\begin{tikzpicture}[scale=0.60]%70
\begin{groupplot}[
    group style={
        group name=my plots,
        group size=2 by 2,
        xlabels at=edge bottom,
        ylabels at=edge left,
        horizontal sep=1cm,
        vertical sep=1.5cm
        },
    ytick={0.0,0.25,0.5,0.75,1.0},
    legend style={at={(0.5,-1.5)},anchor=north, legend columns=4, font=\small},
    xmajorgrids=true,
    ymajorgrids=true,
    cycle list name=exotic,
    grid style=dashed,
    xlabel={\# Labeled Cells},
    ymin=0.0, ymax=1.0,
    ]

\input{charts/features_all/features_flights}
\input{charts/features_all/features_movies}
\input{charts/features_all/features_beers}
\input{charts/features_all/features_address}

\end{groupplot}
\end{tikzpicture}
}
\ref{testLegendfeatures}
\end{center}
	\caption{Comparison of different feature sets for Flights, Beers, and Address.} 
	\label{fig:features}
\end{figure*}

\subsubsection{Evaluation Methodology}
We measure the effectiveness of the methods in detecting potential errors using precision~(P), recall~(R), and $F_1$-score~($F_1$). 
%Additionally, we measure the runtime of each method in seconds. 
We report the labeling effort by the number of data cell labels.

\subsubsection{Our Approach}
\begin{sloppypar}
Per default, we use \system{} with the feature combination of character-level unigrams, metadata, cell value correlation, and error correlation features, and the \emph{Min Certainty} column selection strategy.
We choose an active learning batch size of $k = 10$. Preliminary experiments show that the batch size does not significantly affect performance. 
As the classifier, we choose XGBoost~\cite{chen2016xgboost} because of its robustness against irrelevant features~\cite{friedman2001greedy} (see Section~\ref{sec:model_selection}). 
For the cell value embedding space, we choose $100$ dimensions as proposed by Krishnan et al.~\cite{krishnan2017boostclean}.
Since \system{} is not deterministic, we apply it 10~times and report the average and the standard deviation.
\end{sloppypar}

\subsubsection{Implementation Details}
\label{sec:implementation}
\begin{sloppypar}
%Our current prototype is available at \url{https://github.com/BigDaMa/ExampleDrivenErrorDetection}. 
All experiments were executed on a machine with 14~2.60GHz Intel Xeon E5-2690 CPUs (each two threads), 251GB~RAM, and running Ubuntu~16.04.2.
\end{sloppypar}

\iffalse
flights: 6 * 2376
movies: 17 * 7390
restaurants: 16 * 28787
beers: 10 * 2410
address: 12 * 94306
hospital: 17 * 1000

flights: 116
movies: 300
restaurants: 499
beers: 336
address: 148
hospital: 498

flights: (116 / (6 * 2376)) * 100 = 0.81 %
movies: (300 / (17 * 7390)) * 100 = 0.23 %
restaurants: (499 / (16 * 28787)) * 100 = 0.11 %
beers: (336 / (10 * 2410)) * 100 = 1.39 %
address: (148 / (12 * 94306)) * 100 = 0.013 %
hospital: (418 / (17 * 1000)) * 100 = 2.45 %

(0.81 + 0.23 + 0.11 + 1.39 + 0.013 + 2.93) / 6 = 0.91

flights: (116 / (4 * 2376)) * 100 = 1.22 %
movies: (300 / (10 * 7390)) * 100 = 0.41 %
restaurants: (499 / (10 * 28787)) * 100 = 0.17 %
beers: (336 / (4 * 2410)) * 100 = 3.49 %
address: (148 / (7 * 94306)) * 100 = 0.02 %
hospital: (418 / (17 * 1000)) * 100 = 2.45 %
\fi

\subsection{Effectiveness}
\label{sec:effectivity}

In all our experiments, \system{} achieves state-of-the-art error detection $F_1$-score with only less than 1\% labels.
Figure~\ref{figure:methods_comparison_user_effort} illustrates the $F_1$-score performance of different error detection methods with respect to the number of required labels, if applicable.
% when do we need more labels:
% and on the number of erroneous columns of the dataset and 
The number of required labels depends on the diversity of errors. The higher the error fraction and the less diverse they appear, the fewer labels are needed.
Additionally, Table~\ref{tab:effectivity} reports the best $F_1$-score result of each error detection method from Figure~\ref{figure:methods_comparison_user_effort} to show the corresponding precision and recall. 

% range of labels that are required to outperform all methods 
%Given user labels that range from 0.013\% of all data cells for Address to 1\% of all data cells for Beers, \system{} outperforms all other methods with regard to $F_1$-score.

%For instance, a missing value can be easily detected because the string length is zero and the representation of the text features is the same across all missing values.

% why performs \system{} better than all other tools
The main reason for the superior performance of \system{} is that it can detect a broad set of error types, such as pattern violations, outliers, and constraint violations. Traditional methods, such as NADEEF, dBoost, and NADEEF, cannot reach the performance of \system{} because they are limited to special types of errors. 
For instance, on the Address dataset, \system{} covers at least 96\% of the true positives of each competing method.
%intercolumn correlations
Another advantage of \system{} over conventional methods is that \system{} can effectively exploit inter-column correlations without the help of the user. The Flights dataset requires the error detection method to exploit inter-column correlations to achieve high accuracy. For instance, for a low-quality data source, the data source identifier might correlate with erroneous values in other columns. The user might not even know about these subtle correlations.
%why uncertainty
However, for Movies, \system{}'s results differ significantly across runs. The reason for this uncertainty is that the dataset Movies has a very low error fraction. Since there are very few errors, it is harder for \system{} to find these errors and learn their pattern.

%why outlier detection is bad:
%Unsupervised outlier detection methods, such as the dBoost models Gaussian, Histogram, and Mixture, can only detect outliers. Therefore, these methods fail on datasets with high error fractions. For instance, the erroneous city name~\emph{"SAN"} occurs $5051$~times in the Address dataset. 

%why KATARA is bad:
%KATARA can only find errors for relations that exist in the knowledge base and if these relations exist, they have to be in the same format as the data at hand. KATARA finds a number of matching relations. For instance, for the Movies dataset, KATARA finds relations that apply to the columns \emph{Name}, \emph{Year}, and \emph{Director}. However, these relations do not fit perfectly and result in low precision.

%why NADEEF is bad:
%\mohammad{remove:}In the case of NADEEF, we leverage functional dependencies and simple denial constraints, such as string length, value equality, and data type checks. One reason for inferior performance is that some of the datasets, such as Movies, do not contain meaningful inter-column dependencies. However, in the case of the Beers dataset, four denial constraints can detect all errors perfectly.

Furthermore, the experiments show that both the BoostClean and the ActiveClean features are not expressive enough and therefore achieve lower recall than \system{}.

%why Metadat-Driven Error Detection is bad
Metadata-Driven Error Detection~(MDED) converges faster at the beginning than \system{} because it leverages the results of the configured tools NADEEF, KATARA, and dBoost. However, given enough labels, \system{} always outperforms MDED. Furthermore, MDED fails to reach the performance of NADEEF for Movies due to a large number of false positives returned by KATARA.

% why is BoostClean good and bad
%\mohammad{remove:}The BoostClean feature set performs well if the data is highly redundant and the errors can be detected using cell value correlations, as shown for Flights. However, compared to \system{}, the achieved recall is generally lower because of the missing text representation features. Especially syntactical error detection benefits from text representation features.

% why is ActiveClean doing better in the beginning of and worse else

%\mohammad{remove:}ActiveClean, which leverages bag-of-words features, can cover less fine-granular errors. Therefore, we see a significant lower recall in comparison to \system{}. 

%summary
In summary, we see that, given a small set of labels, \system{} can outperform all other error detection methods while requiring a comparably short runtime.
Our approach requires slightly higher runtime than BoostClean because of the more complex features but is still faster than ActiveClean. On the largest dataset \system{} needs 12 minutes of machine time for the obtained results.
Furthermore, \system{} does not require any tool-specific knowledge.

\begin{figure*}[h!]
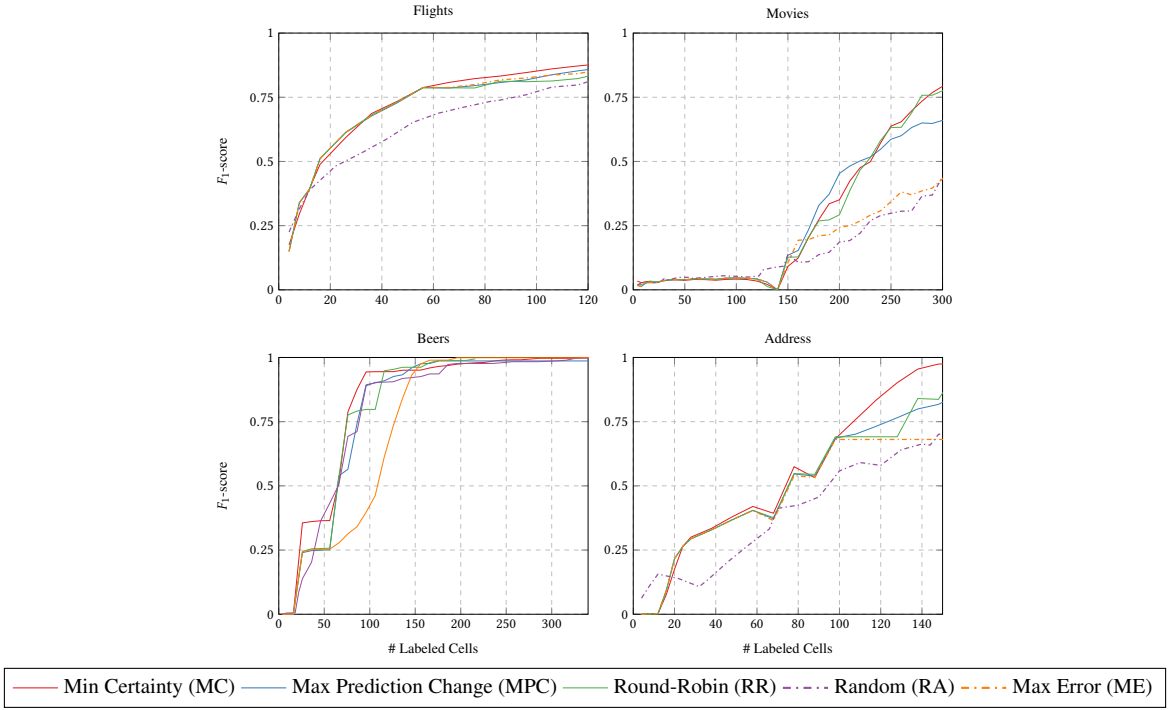

	\centering
	\begin{center}
\trimbox{0cm -0.1cm 0cm 0cm}{
\begin{tikzpicture}[scale=0.60]
\begin{groupplot}[
    group style={
        group name=my plots,
        group size=2 by 2,
        xlabels at=edge bottom,
        ylabels at=edge left,
        horizontal sep=1cm,
        vertical sep=1.5cm
        },
    ytick={0.0,0.25,0.5,0.75,1.0},
    legend style={at={(0.5,-1.5)},anchor=north, legend columns=-1, font=\small},
    xmajorgrids=true,
    ymajorgrids=true,
    cycle list name=mycolorlist,
    grid style=dashed,
    ymin=0.0, ymax=1.0,
    xlabel={\# Labeled Cells}
    ]

\input{charts/order_all/flights}
\input{charts/order_all/movies}
\input{charts/order_all/beers}
\input{charts/order_all/address}

\end{groupplot}
\end{tikzpicture}
}
\ref{testLegendorder}
\end{center}
	\caption{Comparison of different column selection strategies.} 
    \label{fig:order}
\end{figure*}

\subsection{Micro-Benchmark Results}
\label{sec:microbenchmark}

Next, we evaluate different internal decisions that we made for our system.
First, we evaluate the features that we described in Section~\ref{sec:features}. Then, we evaluate the strategies for column selection. Finally, we evaluate different classifiers as the learning model in our system. 

\subsubsection{Feature Representation}
\label{sec:feature_selection_experiment}

Figure~\ref{fig:features} illustrates how the $F_1$-score evolves based on the accumulation of labels for different feature sets.
%General
For all datasets, the holistic feature set~$\rho_{\text{\system{}}}$, which comprises text, metadata, cell value correlation, and error correlation features, reaches the highest $F_1$-score.

%word-level text features
The word-level text features perform poorly because they are not fine-granular enough as described in Section~\ref{sec:text_features}.
%character-level features
Therefore, text features, such as character-level~$\rho_{\text{n-grams}}(n=1)$, perform significantly better on these datasets. 
We see that, for the presented datasets, n-grams of higher order do not improve the $F_1$-score significantly.
%metadata
A surprising outcome is that metadata features~$\rho_{\text{meta}}$ alone can be very powerful to detect a wide range of errors. 
%word2vec
The cell value correlation features~$\rho_{\text{w2v}}$ performance varies significantly across datasets and heavily depends on data redundancy.
%error correlation
For datasets that contain error correlations, such as Flights, the additional error correlation features accelerate convergence.

In summary, the \system{} features represent the best trade-off among $F_1$-score, feature generation runtime, training time. 
%Additionally, we discussed how to leverage these features to explain both the trained models and the error causes.

\subsubsection{Column Selection Strategy}
\label{subsec:orderstrategy}

%We evaluate different sampling strategies for choosing the next column for labeling. 
The goal of the sampling strategies is to induce the steepest learning curves to achieve a high $F_1$-score as fast as possible. 
Figure~\ref{fig:order} illustrates how the column selection strategies proposed in Section~\ref{sec:order} perform on three exemplary datasets.

%Random
As expected, the Random strategy performs poorly because it does not consider model convergence.
%Min Certainty 
In contrast to the Random strategy, we see that the Min Certainty strategy achieves the best $F_1$-score across all datasets. 
However, the benefit of using the Min Certainty strategy is small in the case of Movies. The Min Certainty strategy performs best if there is a large number of erroneous columns that differ significantly in their error complexity and therefore convergence. Movies has similar error fractions across columns.
%Address
In the Address dataset, error complexity across columns varies significantly because the data types and error fractions vary across columns. 
Both the Max Prediction Change and the Max Error strategy converge slower than the Min Certainty strategy and potentially get stuck in local optima because the predictions or the cross-validation error do not change in spite of learning progress, as we can see on Address.
The Round-Robin strategy performs well if the model convergence for all erroneous columns is similar.

%summary
In summary, because of the fast convergence and the high stable $F_1$-score across datasets, we choose to use the Min Certainty~(MC) sampling strategy for \system{}.

\begin{figure}[h]
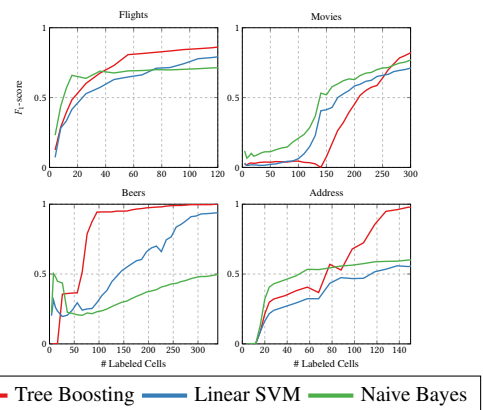

	\centering
	\begin{center}
\scalebox{0.48}{
\trimbox{0cm -0.1cm 0cm 0cm}{
\begin{tikzpicture}[scale=0.68, font=\LARGE]
\begin{groupplot}[
    group style={
        group name=my plots,
        group size=2 by 2,
        xlabels at=edge bottom,
        ylabels at=edge left,
        horizontal sep=1cm,
        vertical sep=1.5cm
        },
    ytick={0.0,0.5,1.0},
    legend style={at={(0.5,-1.5)},anchor=north, legend columns=-1, font=\small},
    xmajorgrids=true,
    ymajorgrids=true,
    cycle list name=mycolorlist,
    grid style=dashed,
    %label style={font=\LARGE},
    %tick label style={font=\LARGE},
    xlabel={\# Labeled Cells},
    ymin=0.0, ymax=1.0,
    xmin=0.0
    ]

\input{charts/model_all/flights}
\input{charts/model_all/movies}
\input{charts/model_all/beers}
\input{charts/model_all/address}

\end{groupplot}
\end{tikzpicture}
}
}
\ref{testLegendmodels}
\end{center}	
	\caption{Comparison of different classification models.} 
	\label{fig:model}
\end{figure}

\subsubsection{Model selection}
\label{sec:model_selection}
\begin{sloppypar}
We evaluate the effect of different classifiers on our system. These classifiers are the tree boosting algorithm XGBoost~\cite{chen2016xgboost}, multinomial Naive Bayes~\cite{mccallum1998comparison}, and a support vector machine~(SVM) using a linear kernel~\cite{cortes1995support}, which have been used for text classification use cases~\cite{tong2001support, chen2016xgboost, mccallum1998comparison}. Figure~\ref{fig:model} illustrates the results of this evaluation. 
Although Naive Bayes and SVM perform better on very small training sets, XGBoost achieves the highest $F_1$-score within the fastest convergence on all three datasets. 
The main reason is that trees are robust against irrelevant features and can handle missing values~\cite{friedman2001greedy}. 
%For instance, for the Hospital dataset, only one out of $2165$ features is relevant to detect the errors, namely whether or not the character 'x' is present. 
SVMs perform poorly in the case of many irrelevant features~\cite{weston2001feature}. Furthermore, functional dependencies in the datasets violate Naive Bayes' independence assumption~\cite{rennie2003tackling}. Additionally, class imbalance negatively affects the performance of Naive Bayes~\cite{rennie2003tackling}.

In summary, XGBoost appears to be more robust than the competitors for \system{} because it can successfully ignore irrelevant noisy features. 
\end{sloppypar}

\section{Related Work}
\label{sec:relatedwork}
Our approach combines multiple areas of research. %We base our work on research in error detection, feature engineering, and active learning. 
First, we discuss existing research on traditional error detection. 
Then, we compare our feature representation with representations that have been used in other machine learning-based error detection methods.
Finally, we discuss existing work on active learning with a focus on data cleaning applications.

\subsection{Traditional Error Detection}
Error detection has received extensive attention in the information management community~\cite{abedjan2016detecting, chu2015katara, dallachiesa2013nadeef, pit2016outlier}. 
Different error detection methods are typically tailored towards different types of data errors. 
The two main categories of methods are quantitative or qualitative error detection~\cite{chu2016data}. 
Abedjan et al. further identify four subcategories: rule violation detection, pattern violation detection, outlier detection, and duplicate conflict resolution~\cite{abedjan2016detecting}. 
Rule-based systems, such as NADEEF~\cite{dallachiesa2013nadeef}, can compile business rules and data patterns as denial constraints~\cite{chu2013holistic} to uncover errors. 
%Pattern enforcement and transformation methods leverage syntactic, e.g., OpenRefine~\cite{verborgh2013using} and Data Wrangler~\cite{kandel2012enterprise}, 
Other systems such as KATARA~\cite{chu2015katara} use external reference data to uncover semantic errors. 
Outlier detection methods, such as dBoost~\cite{pit2016outlier}, consider values as errors if they deviate from the given norm.
%Moreover, record linkage and de-duplication methods, such as Data Tamer~\cite{stonebraker2013data}, find duplicate data records. For duplicates, conflicting values for the same attribute indicate errors.
All these methods require the user to provide configuration parameters, whereas \system{} does not require any configuration. 

\subsection{Machine Learning-Based Error Detection}
Machine learning-based error detection methods leverage a broad variety of feature representations ranging from simple rules and patterns to sophisticated language models~\cite{pit2016outlier,krishnan2017boostclean}.
Pit-Claudel et al.~\cite{pit2016outlier} apply tuple expansion based on pre-defined patterns. Simple transformation rules, such as extracting numbers from a string attribute, are initial steps to find errors in heterogeneous data~\cite{pit2016outlier,visengeriyeva2018metadata,krishnan2017boostclean}. 
Krishnan et al.~\cite{krishnan2016activeclean} use the TF-IDF representation of data values to train a model that classifies whether a tuple is dirty or not. We show in our experiments that a character-level representation achieves higher accuracy than a word-level representation. In contrast to ActiveClean's tuple-wise classification~\cite{krishnan2016activeclean}, \system{} can identify the actual erroneous values inside the tuples. Furthermore, BoostClean~\cite{krishnan2017boostclean} leverages the \emph{Word2vec} approach~\cite{mikolov2013distributed} to model cell value correlations.
Using language models, such as bag-of-words and n-grams~\cite{cavnar1994n}, is common for text classification~\cite{joachims1998text, mccallum1998comparison}.
However, for the use case of error detection, we showed with an in-depth analysis that the language model can be advanced by other features, such as column concatenation, metadata, and error correlation features.

Recently, machine learning has been leveraged to improve data cleaning with respect to data repair~\cite{yakout2011guided,rekatsinas2017holoclean,yakout2013don}. In this paper, we focus on the error detection task and not the data repairing step. Accordingly, research on data repair considers error detection as a black box~\cite{rekatsinas2017holoclean}.

\subsection{Active Learning}

Active learning~\cite{settles2010active} has been used in various data cleaning tasks, such as duplicate detection~\cite{firmani2016online, gokhale2014corleone}, outlier detection for numerical data~\cite{abe2006outlier, steinwart2005classification}, and data repair~\cite{krishnan2016activeclean,yakout2011guided}.
To the best of our knowledge, \system{} is the first approach that leverages active learning to address the task of error detection for relational data with heterogeneous data types without user configuration. 

Both ActiveClean and BoostClean do not apply active learning~\cite{krishnan2016activeclean,krishnan2017boostclean}. Both assume that the user trains an application-specific machine learning model on the dataset in question. Therefore, the user has to provide labels for the corresponding machine learning task for all tuples of the dataset. In contrast, \system{} is a stand-alone cell-wise error detection method with its own active learning strategy that does not require a use case.
In general, we can interpret error detection as a text classification task~\cite{tong2001support}. However, in order to exploit the priors of relational data, we define it as a two-dimensional \emph{multi-classifier} active learning problem. This problem does not coincide with the two-dimensional \emph{multi-label} active learning problem~\cite{qi2008two}. To select the most beneficial column for labeling, we leverage heuristics that were proposed as stopping criteria for active learning~\cite{bloodgood2009method, zhu2007active}.

\section{Conclusions}
\label{sec:conclusions}
\begin{sloppypar}
We introduced a new example-driven approach to error detection that relies on active learning. Our approach does not require any parameter configuration. We empirically showed that \system{} achieves state-of-the-art $F_1$-score in comparison to other error detection methods while observing comparably low user effort. In general, our approach requires significantly fewer labels than the competing learning-based approaches to achieve the same $F_1$-score.
\system{} can detect a large variety of error types ranging from outliers over pattern violations to constraint violations and has an optimized active learning strategy. Additionally, \system{} can exploit hidden inter-column correlations to uncover complex error patterns that the user was unaware of upfront.
%The combination of the character-level unigram, metadata, cell value correlation, and error correlation features allows \system{} to model relational data effectively. Furthermore, this combination enables fast feature generation and fast training. 

%Our insights motivate several research directions. 
%In particular, the application of novel reinforcement learning and meta-learning techniques might be interesting directions for error detection. 
%Moreover, the active learning cold start problem requires more research. 
As a future direction, it would be interesting to extend our method to data repair or to allow for a crowdsourcing scenario with multiple noisy, imperfect users~\cite{sheng2008get}.
%Finally, we want to extend our method to data repair.
\end{sloppypar}
\section*{Acknowledgments}

The authors would like to thank Rekatsinas et al. for providing us with their datasets Flights and Hospital and the corresponding ground truth \cite{rekatsinas2017holoclean}. 
\begin{sloppypar}
Furthermore, we like to thank Jens Meiners, Stephan Alaniz, Mihail Bogojeski, Sebastian Schelter, Jan Laermann, Felix Bie{\ss}mann, Kevin Kepp, Martin Kiefer, Tobias Fuchs, Larysa Visengeriyeva, Dennis Schmidt, Alexander Renz-Wieland, Mahdi Esmailoghli, Sergey Redyuk, Tilmann Rabl, and G\'abor~E.~G\'evay for discussions and feedback.

Finally, this work was supported in the context of the DAYSTREAM project funded by the German Federal Ministry of Transport and Digital Infrastructure~(19F2031A).
\end{sloppypar}

%
% The next two lines define the bibliography style to be used, and the bibliography file.
\bibliographystyle{ACM-Reference-Format}
\bibliography{abbreviations,sigproc_short}

% 
% If your work has an appendix, this is the place to put it.
%\appendix

%\section{Research Methods}

\end{document}